\documentclass{article}

\usepackage[nonatbib, preprint]{neurips_2022}

\usepackage[style=numeric-comp]{biblatex}
\usepackage[utf8]{inputenc} % allow utf-8 input
\usepackage[T1]{fontenc}    % use 8-bit T1 fonts
\usepackage{hyperref}       % hyperlinks
\usepackage{url}            % simple URL typesetting
\usepackage{booktabs}       % professional-quality tables
\usepackage{amsfonts}       % blackboard math symbols
\usepackage{nicefrac}       % compact symbols for 1/2, etc.
\usepackage{microtype}      % microtypography
\usepackage{xcolor}         % colors
\usepackage{graphicx}
\usepackage{amsmath}
\usepackage{float}

\title{Spatio-temporally separable non-linear latent factor learning: an application to somatomotor cortex fMRI data}

\author{%
  Eloy Geenjaar \\
  TReNDS Center \\
  Georgia Institute of Technology\\
  Atlanta, GA \\
  \texttt{egeenjaar@gatech.edu} \\
  \And 
  Amrit Kashyap \\
  Department of Brain Simulation \\
  Charite University Hospital \\
  Berlin, Germany \\
  \And
  Noah Lewis \\
  TReNDS Center \\
  Georgia Institute of Technology \\
  Atlanta, GA\\
  \And 
  Robyn Miller \\
  TReNDS Center \\
  Georgia State University \\
  Atlanta, GA \\
  \And
  Vince Calhoun \\
  TReNDS Center \\
  Georgia Institute of Technology \\
  Atlanta, GA \\
}

\begin{document}

\maketitle

\begin{abstract}
Functional magnetic resonance imaging (fMRI) data contain complex spatiotemporal dynamics, thus researchers have developed approaches that reduce the dimensionality of the signal while extracting relevant and interpretable dynamics. Recently, the feasibility of latent factor analysis, which can identify the lower-dimensional trajectory of neuronal population activations, has been demonstrated on both spiking and calcium imaging data. In this work, we propose a new framework inspired by latent factor analysis and apply it to functional whole-brain data from the human somatomotor cortex. Models of fMRI data that can perform whole-brain discovery of dynamical latent factors are understudied. The benefits of approaches such as linear independent component analysis models have been widely appreciated, however, nonlinear extensions of these models present challenges in terms of identification. Deep learning methods provide a way forward, but new methods for efficient spatial weight-sharing are critical to deal with the high dimensionality of the data and the presence of noise. Our approach generalizes weight sharing to non-Euclidean neuroimaging data by first performing spectral clustering based on the structural and functional similarity between voxels. The spectral clusters and their assignments can then be used as patches in an adapted multi-layer perceptron (MLP)-mixer model to share parameters among input points. To encourage temporally independent latent factors, we use an additional total correlation term in the loss. Our approach is evaluated on data with multiple motor sub-tasks to assess whether the model captures disentangled latent factors that correspond to each sub-task. Then, to assess the latent factors we find further, we compare the spatial location of each latent factor to the motor homunculus. Finally, we show that our approach captures task effects better than the current gold standard of source signal separation, independent component analysis (ICA). 
\end{abstract}

\section{Introduction}
Functional magnetic resonance imaging (fMRI) is an important and widely used imaging method to study the whole-brain dynamics of the human brain. Although it does not directly capture neuronal activity, it can serve as a proxy for measuring neuronal activity non-invasively at high spatial resolutions. Clinicians and researchers have had issues interpreting the signal, however, due to the signal's high dimensionality, poor temporal resolution, and multiple sources of noise leading to a low signal-to-noise ratio. Both to understand the signal itself better and to move towards potentially clinically relevant information, researchers have focused on developing methods that summarize the signal across spatial and temporal scales \cite{descombes1998spatio, woolrich2004fully}. Pre-defined Atlases are also a popular tool to average and increase the signal-to-noise ratio in brain regions of interest (ROI). Averaging can be misleading because spatial regions can have multiple distinct timecourses that overlap within each region, which has led researchers to tools such as independent component analysis (ICA), that decompose the signal into multiple temporal trajectories with corresponding spatial sources \cite{mckeown1998independent, beckmann2005tensorial, calhoun2006unmixing}.
A promising alternative is emerging with respect to the characterization of neuronal population dynamics using fully-differentiable data-driven approaches. These approaches can scale to large neurological data easily, as well as allow for individualized trainable models. One example of such a technique is latent factor analysis via autoencoders \cite{yu2008gaussian, everett2013introduction}. Classically, latent factor analysis for fMRI data is done with some form of matrix factorization, such as principal component analysis \cite{thomas2002noise}, ICA \cite{mckeown1998independent, beckmann2005tensorial, calhoun2006unmixing}, or dictionary learning \cite{lee2010data}. Recently these matrix factorizations have been extended to tensor factorizations/analysis \cite{ma2016spatio}, restricted Boltzmann machines (RBM)s \cite{hjelm2014restricted}, and static autoencoders \cite{kim2021representation, geenjaar2021variational}. In the field of neuronal populations, however, a recent approach finds latent factors using a recurrent autoencoder \cite{pandarinath2018inferring}. Although with different interpretations and under different constraints, low-dimensional latent dynamics underlying the fMRI signal also likely exist. As such, an adapted application of latent factor analysis using recurrent autoencoders is a direction that can help alleviate some of the issues commonly associated with fMRI data. Although this approach does not directly tackle the fact that fMRI data is a proxy for neuronal activity, with careful interpretation, we can still make inferences about functional whole-brain dynamics. Moreover, deep learning approaches allow for the inclusion of relevant constraints, such as geodesic distances or known functional constraints that can further help constrain the solution space to relevant brain dynamics. In this work, we try to find dynamic latent factors underlying the fMRI signal in the context of motor task activations. In this paradigm, the ground truth factors are the motor regions and their associated activations, and the motor homunculus is well-documented across species. For a null model, we utilize the current gold standard of decomposing spatio-temporal fMRI signals, ICA. In addition, we explore the inclusion of constraints to the dynamics by using weight sharing based on hemispheric symmetry, geodesic distances, as well as prior functional activation. We show that our approach combined with these inductive biases better captures task effects than linear ICA when decomposing neural activity. Some recent work in this direction also identifies meaningful non-linear dynamical systems from fMRI task data, but they use task information as input to their model \cite{koppe2019identifying}, whereas we do not. Other related work \cite{gao2020non} shows that there exists a non-linear manifold for all tasks in the HCP dataset, but does not look at any specific factors and use ROIs to decrease the dimensionality of the data.

\section{Method}
The methods are organized by first explaining the experimental setup of the motor task fMRI data, as well as the relevant biological data used. The ground truth activations are based on the generalized linear model hemodynamic responses derived from SPM \cite{penny2011statistical} for each of the sub-motor tasks. We then describe and explain our method. The subsequent sections detail our novel weight sharing method to reduce the number of parameters and more directly incorporate neuroscientific inductive biases. Subsequently, we explain the temporal independence factor we include in our objective function and how we evaluate the temporal factors. The comparisons to null models using ICA are established in the final sections.  

\paragraph{Biological data} 
\label{sec:biological-data}
The data we use in this work are cortical surface timeseries from the open-access, under data usage terms, HCP-1200 dataset \cite{van2013wu}, for all subjects with cortical surface timeseries data (1181). The data is registered using multimodal surface registration (MSM) \cite{robinson2014msm, robinson2018multimodal}, and surfaces are constructed using Freesurfer \cite{glasser2013minimal, fischl2012freesurfer}. Then, the voxels corresponding to the somatomotor region are extracted using the Yeo-7 atlas \cite{yeo2011organization}. Each subject's timeseries is band-pass filtered independently ($0.01-0.15$Hz) and then linearly detrended using the Nilearn package \cite{abraham2014machine}. The cortical surface is represented as a set of voxels ($\mathcal{V}$) and each voxel has a blood-oxygen-level-dependent (BOLD) value associated with it at each timestep ($t$). Each timeseries, for each subject, is mapped into a group space, which means that each voxel represents roughly the same location in the brain. This also means that some deformation is not only introduced in the process of obtaining the surface voxels but also during the registration of the timeseries into group space. The locations of the voxels in this work are based on the group-based pial surface, which is the boundary between gray and white matter in the brain. The surface is thus essentially a graph with a fixed structure, and only the values associated with the voxels change over time. Although cortical surface data has previously been mapped to a sphere and then been mapped to a $2D$ image using polar coordinates \cite{kim2021representation}, in this work we view the location of each vertex as a graph to retain as much distance information as possible. 

\paragraph{Metrics and experimental setting} 
The data consists of 5 motor tasks, left hand, right hand, left foot, right foot, and tongue movements, where the subject moves the respective limbs after hearing an auditory cue. Since the regions associated with these movements are well defined based on the motor homunculus as well as the timing of these events is known due to the event-based scientific paradigm, the ground truth of the spatio-temporal signal is well established in these tasks. Therefore we evaluate our model based on the following metrics: 1) its ability to reproduce the correct spatial maps observed during motor activation, and 2) the exact temporal dynamics associated with the activation during these tasks. The dataset is randomly shuffled and then divided into a training set ($70\%$), a validation set ($10\%$), and a test set ($20\%$) to make sure it generalizes beyond the training data.

The tasks are assumed to last for $12$ seconds, and each factor's timeseries is convolved with SPM's simulated hemodynamic response based on the block design of each sub-task. After obtaining the average temporal timeseries for the unseen test set, we find the factors that have the highest absolute average correlation with each sub-task. We then take the average over those absolute correlations to measure how well the model can learn some ground-truth underlying factors in the dataset. Knowing the model finds underlying factors in the dataset opens up using this model for resting-state data, where underlying factors are often less apparent and no ground truth exists.

\paragraph{Sequential variational autoencoder}
\label{ref:method:sequential-autoencoder}
Sequential autoencoders were developed to learn and model temporal dynamics efficiently. From an information-theoretic perspective, they bottleneck the information and assume that only the most important information is retained in the latent space. As such, sequential autoencoders have been used in a variety of different problems in order to model temporal datasets such as speech processing \cite{graves2013speech}, to compress high dimensional neuronal population data \cite{keshtkaran2021large}, as well as model fMRI dynamics \cite{kashyap2020brain}.

\begin{figure}[H]
    \centering
    \includegraphics[width=\textwidth]{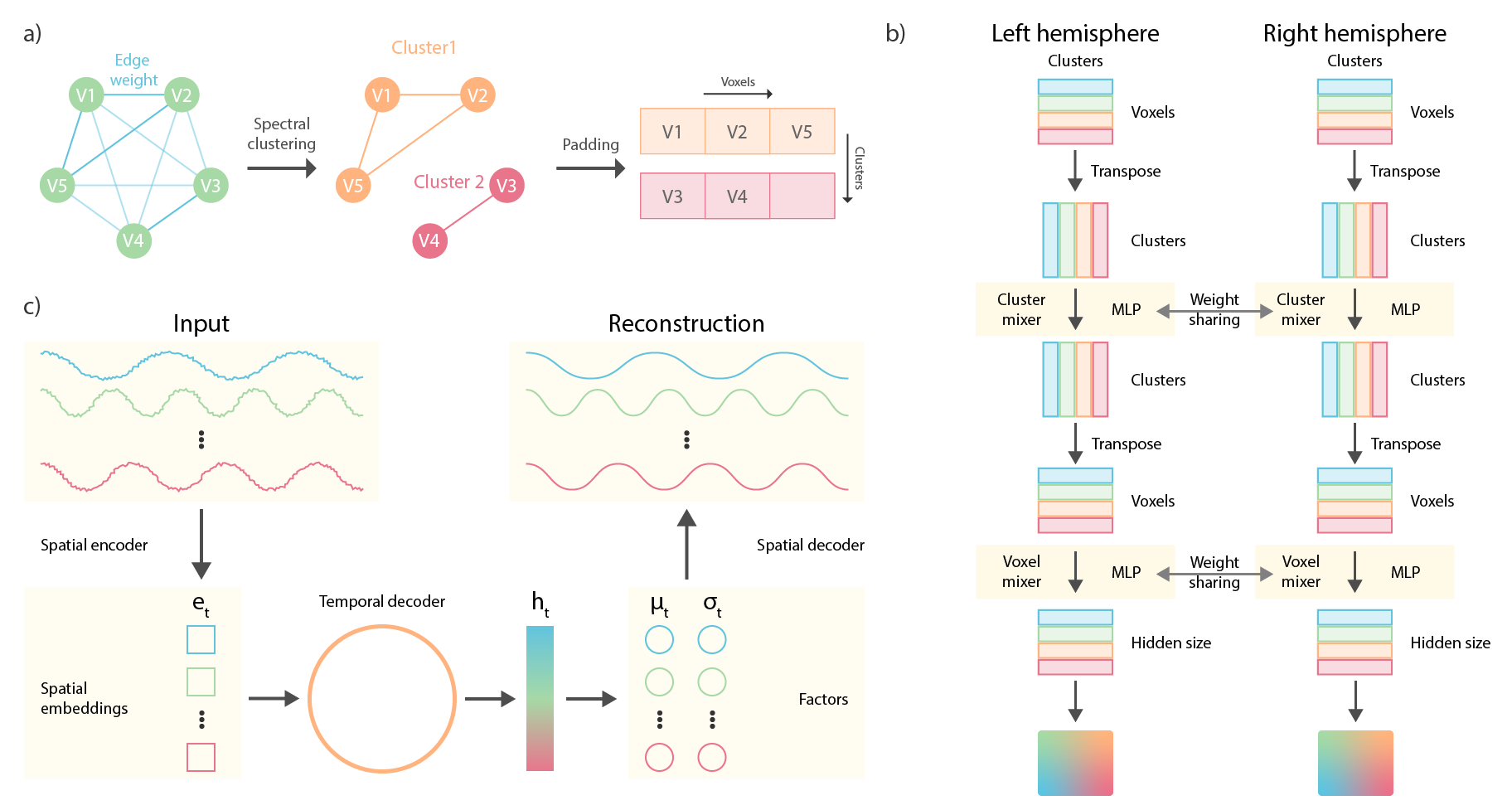}
    \caption{The top left (a) of the figure pictorially illustrates how spectral clustering works. Top right (b) shows the mixer layer we use in both the spatial encoder and decoder. The MLP consists of two linear layers, with an ELU activation on the first layer. The bottom left subfigure (c) shows the full dynamic model.}
    \label{fig:model}
\end{figure}

The sequential variational autoencoder in this methodology consists of a gated recurrent unit (GRU)\cite{cho2014properties} and a linear layer. The GRU obtains as input the embeddings from the spatial encoder $e_{t}$ and outputs its hidden state at each timestep. These hidden states are used to parameterize the mean and standard deviation of the Gaussian distributions at each timestep, see Figure \ref{fig:model}c. The distributions are referred to as the factors $f_t$ in this work. The reason we model the factors as distributions is that the loss function of variational autoencoders \cite{kingma2013auto} has been shown to encourage disentanglement of the separate factors in each distribution \cite{graves2008novel, higgins2016beta, burgess2018understanding, higgins2022symmetry}. 

Formally, the problem consists of a dataset $\{\mathbf{x}^{(1)}, \mathbf{x}^{(2)}, ..., \mathbf{x}^{(\text{N})}\} \in \mathcal{D}$, where each $\mathbf{x}^{(\text{i})}$ is made up of $T$ timesteps $\mathbf{x}^{(\text{i})} = \{\mathbf{x}^{(\text{i})}_{1}, \mathbf{x}^{i}_{2}, ..., \mathbf{x}^{\text{(i)}}_{\text{T}} \}$. Each timestep for a subject $\mathbf{x}^{\text{(i)}}_{\text{t}}$ are the blood-oxygen-level dependent (BOLD) values for each input voxels at that time. The model proposed in this work is based on a variational autoencoder (VAE) \cite{kingma2013auto}, which learns both a generative $p_{\mathbf{\theta}}(\mathbf{x} | \mathbf{z})$ and a variational approximation $q_{\mathbf{\phi}}(\mathbf{z}  | \mathbf{x})$ of the true posterior. VAEs are optimized using the evidence lower-bound (ELBO) on the expected marginal log-likelihood of $\mathbf{x}$, a more in-depth explanation of the ELBO is provided in previous work \cite{kingma2013auto}. In our case we obtain a latent variable for each subject $\mathbf{z}^{(i)}$ and for each timestep $\mathbf{z}^{(i)} = \{ \mathbf{z}^{(i)}_{1}, \mathbf{z}^{(i)}_{2}, ..., \mathbf{z}^{(i)}_{T} \}$. In our work we assume the prior for the variational estimation to be a zero-mean, unit-norm diagonal multivariate Gaussian distribution $p(\mathbf{z})$. The resulting ELBO for our problem setting is as follows.
\begin{equation}
    \label{eq:elbo}
    \mathcal{L}(\mathbf{\theta}, \mathbf{\phi}; x^{\text{(i)}}_{\text{t}}) := 
    -\text{D}_{\text{KL}}
    \left(q_{\mathbf{\phi}}(\mathbf{z}^{\text{(i)}}_{\text{t}}  | \mathbf{x}^{\text{(i)}}_{\leq \text{t}}) || 
    p(\mathbf{z})
    \right) + \mathbb{E}_{q_{\mathbf{\phi}}(\mathbf{z}^{\text{(i)}}_{\leq \text{t}}  | \mathbf{x}^{\text{(i)}}_{\text{t}})} \left[\log p_{\mathbf{\theta}}(
    \mathbf{x}^{\text{(i)}}_{\text{t}} | \mathbf{z}^{\text{(i)}}_{\text{t}})
    \right]
\end{equation}

The two terms can be seen as an encoder $q_{\mathbf{\phi}}(\mathbf{z}^{\text{(i)}}_{\text{t}}  | \mathbf{x}^{\text{(i)}}_{\leq \text{t}})$ and a decoder $p_{\mathbf{\theta}}( \mathbf{x}^{\text{(i)}}_{\text{t}} | \mathbf{z}^{\text{(i)}}_{\text{t}})$, both parameterized by separate neural networks. If the variance of the input data is assumed to be constant, then optimizing the log-likelihood of the decoder is the same as optimizing the mean-squared error between the input data and the reconstructed data from the decoder. In this case, the parameters of the encoder $\mathbf{\theta}$ correspond to the spatial encoder and temporal decoder, whereas the generative parameters $\mathbf{\phi}$ correspond to the spatial decoder and distribution parameters (mean and standard deviations). The optimization of this lower bound is in our case done by taking the mean over the dimensions of the distribution and timesteps for the KL-divergence term. We take the sum over the mean-squared error between the reconstructed and true timesteps within a subject but take the mean over the number of input dimensions.

\paragraph{Spectral clustering}
\label{method:spectral-clustering}

To perform spatial reduction of the data and to introduce non-linearities that help the model learn more complex factors, we use a spatial encoder and decoder. Since the neural populations have distinct spatial as well as temporal similarities, spectral clustering is used to group the data based on spatial and temporal similarities. To induce the bias that neuronal circuits that lie closer together on the cortical surface are more likely to be related to each other, we perform spectral clustering with geodesic distances \cite{ng2001spectral, pedregosa2011scikit}. To induce the inductive bias that similarly activated voxels likely communicate with each other more, we also cluster based on the correlations (functional) between the voxels, see Figure \ref{fig:model}a. The geodesic distance is a more realistic distance metric in terms of structural closeness between neuronal populations for MRI data because it takes the gyri and sulci (folds) of the cortical surface into account. Furthermore, we cluster between voxels that have similar temporal activation based on the temporal correlation between them. The spectral clustering then essentially leaves us with 'graph patches' and we can therefore make use of recent work that has explored the use of patches to perform classification \cite{tolstikhin2021mlp}. 

To cluster the vertices based on their structural and functional distances, we first create an adjacency matrix $\mathbf{A}_{s}$ and $\mathbf{A}_{f}$ for each, where each row and column corresponds to a voxel ($v \in \mathcal{V}$). The geodesic distances for each voxel ($e \in \mathcal{E}$) within a hemisphere are calculated using the HCP connectome workbench \cite{van2013wu} and the functional connections are calculated with a Pearson correlation for all subjects in the training and validation set. The clustering was performed using scikit-learn \cite{pedregosa2011scikit}, but for readers that are unfamiliar with spectral clustering, it generally works as follows. First, we create the adjacency matrix ($\mathbf{A}_{s}$ or $\mathbf{A}_{f}$) and normalize it between $0$ and $1$. We assume a fully-connected graph within each hemisphere, so the degree matrix of the graph is the total number of vertices on the diagonal. Then, the graph Laplacian of the graph is computed as $\mathbf{L}_{s} \; = \; \mathbf{D}_{s} \; - \; \mathbf{A}_{s}$ and similarly for the functional adjacency matrix. Second, the graph Laplacian is decomposed using its eigendecomposition and only the bottom $k$ smallest eigenvalues are used, the others are discarded. The $k$ smallest eigenvalues each correspond to a cluster (eigenvector) of the graph Laplacian. Given that they are based on the structural and functional proximity of each voxel to another, each cluster is interpreted as a patch for the rest of the paper. This allows us to induce specific biases, such as translational invariance, and more importantly parameter sharing, since the dimensionality of the data is large. To improve the parameter sharing of the spatial encoder and decoder even more, the encoder and decoder are shared across each hemisphere. This inductive bias does not imply that the two hemispheres are perfectly symmetric, but rather that we expect the features we need to extract from each hemisphere is similar. The number of clusters we use in this work per hemisphere for both the structural and functional inductive bias is $128$.

\paragraph{Mixer layer}
\label{ref:method:encoder-decoder}
The architecture of the spatial encoder and decoder are based roughly on the MLP-mixer architecture \cite{tolstikhin2021mlp}, without the layer normalization and residual connections. As mentioned before, the spectral clusters are interpreted as patches, and the size of each cluster is used as the size of each patch. Since the sizes of the clusters are not exactly equal, we take the maximum cluster size and pad all the other clusters with zeros up to the maximum size. To make sure this does not result in large pads, we assess the homogeneity of the cluster sizes, which based on visual inspection, are fairly equal. Each mixer layer consists of two multi-layer perceptrons (MLPs), where each MLP consists of two linear layers with an ELU activation \cite{clevert2015fast}. A pictorial view of the mixer layer is shown in Figure \ref{fig:model}b. First, the input is permuted to perform mixing over the input size of the data using the first MLP. The output of the first MLP is again permuted to perform patch-mixing with the second MLP. These blocks are repeated three times, both in the spatial encoder and decoder of our model (see Figure \ref{fig:model}c). In the spatial encoder, the input size is reduced with each subsequent layer, whereas in the spatial decoder, the size is increased with each subsequent layer. The spatial encoder consists of a linear embedding layer, one for each hemisphere separately, that is shared among clusters and maps each cluster to a $128$-dimensional embedding. Each subsequent layer is shared between the two hemispheres and both hemisphere embeddings are passed through 3 mixer layers with hidden sizes $(64, 32, 16, 8, 4, 1)$ for each layer in the MLPs. The output of each hemisphere is then concatenated and mapped from a $256$-dimensional embedding to a $128$-dimensional embedding using a linear layer and ELU activation function \cite{clevert2015fast}. This embedding is used as the input for the temporal decoder, see Figure \ref{fig:model}c. The spatial decoder is exactly symmetric with the spatial encoder, the factors are first mapped to a $256$-dimensional embedding and split into the left and right hemispheres. The left and right hemisphere embeddings are then increased in size spatially using three mixer layers with hidden sizes $(4, 8, 16, 32, 64, 128)$ for each layer in the MLPs, and a separate linear layer for each hemisphere as the final layer in the spatial decoder, with a Tanh activation function. Note that the number of clusters is kept constant throughout the spatial encoder and decoder at $128$, only the spatial size at each layer is increased or decreased.

\paragraph{Connection to non-linear ICA}
Variational autoencoders can under some conditions also perform non-linear ICA with identifiability guarantees \cite{khemakhem2020variational, hyvarinen2019nonlinear}. The way the latent factors are modelled in this work can be considered such a condition, where the additionaly observed variable $\mathbf{u}^{(i)}$ are the previous timesteps in the timeseries. Namely, each factor is a conditional distribution $p_{\mathbf{\theta}}(\mathbf{z}^{(i)}_{t} |  \mathbf{x}^{(i)}_{\leq t})$, where $\mathbf{\theta}$ correspond to the spatial and temporal encoder, and the temporal decoder. This can be rewritten as $p_{\mathbf{\theta}}(\mathbf{z}^{(i)}_{t} | \textbf{x}^{(i)}_{t}, \mathbf{u}^{i)})$. This is the same formulation for the encoder as in the unifying framework for variational autoencoders and non-linear ICA \cite{khemakhem2020variational}.

\paragraph{Temporal independence}
\label{sec:method:temporal-indepence}
The KL-divergence term in the ELBO effectively acts as a regularization term and in previous works \cite{zhao2017infovae,chen2018isolating} has been shown to be equivalent to the following decomposition with an expectation over the dataset $\mathbb{E}_{\mathcal{D}}$.
\begin{align}
    \label{eq:tc-decomposition}
    \mathbb{E}_{\mathcal{D}} \left[\text{D}_{\text{KL}}
    \left(q_{\mathbf{\phi}}(
        \mathbf{z}|\mathbf{x})
        || p(\mathbf{z})
    \right) \right] &=
    \text{D}_{\text{KL}}
        \left(q_{\mathbf{\phi}}(\mathbf{z}, \mathbf{x} )
        || q_{\mathbf{\phi}}(\mathbf{z})p(\mathbf{x}) \right) \tag*{(Index-Code MI)} \\
    & + \text{D}_\text{KL} (
    q_{\mathbf{\phi}}(\mathbf{z})
    || \prod_{j} q_{\mathbf{\phi}}(\mathbf{z}_{\text{j}}) ) \tag{Total correlation} \\
    & + \sum_{\text{j}} \text{D}_{\text{KL}} \left(
    q_{\mathbf{\phi}}(\mathbf{z}_{\text{j}})
    ||p(\mathbf{z}_{\text{j}}) \right) \tag{Dimension-wise KL}
     \\
\end{align}
Where $q_{\mathbf{\phi}}(\mathbf{z}) = \sum_{i=1}^{N} q(\mathbf{z}|\mathbf{x}^{\text{(i)}})p(\mathbf{x}^{\text{(i)}})$ is the aggregated posterior and $\mathbf{z}^{\text{(i)}}_{j}$ is the $j^{\text{th}}$ dimension of the latent factor. Note that with $p(\mathbf{x}^{\text{(i)}}$ we refer to the probability that the sample is chosen as a training sample, which is $\frac{1}{N}$. The authors \cite{chen2018isolating} propose to use minibatch-weighted sampling to get a na\"{i}ve Monte Carlo estimation of aggregated posterior to compute the total correlation (TC) term and identify the TC term as important to learn disentangled factors. The TC measures the dependency among a set of random variables, in this case, the dimensions of the latent factors. In our case, however, we specifically want to minimize the dependency between the factors over time. Thus, instead of estimating the aggregated posterior using samples in the batch, we estimate it over the timesteps for each subject and take the average TC over the batch. We add the TC term to the ELBO and due to the non-negativity of the KL-divergence, this is still a lower bound.
\begin{align}
    \label{eq:final-bound}
    \mathcal{L}(\mathbf{\theta}, \mathbf{\phi}; x^{\text{(i)}}_{\text{t}}) &:= 
    -\text{D}_{\text{KL}}
    \left(q_{\mathbf{\phi}}(\mathbf{z}^{\text{(i)}}_{\text{t}}  | \mathbf{x}^{\text{(i)}}_{\leq \text{t}}) || 
    p(\mathbf{z})
    \right) + \mathbb{E}_{q_{\mathbf{\phi}}(\mathbf{z}^{\text{(i)}}_{\text{t}}  | \mathbf{x}^{\text{(i)}}_{\leq \text{t}})} \left[\log p_{\mathbf{\theta}}(
    \mathbf{x}^{\text{(i)}}_{\text{t}} | \mathbf{z}^{\text{(i)}}_{\text{t}})
    \right] \notag \\
    & - \beta \; \text{D}_{\text{KL}}(q_{\mathbf{\phi}}(\mathbf{z}^{\text{(i)}}_{\text{t}}
    || \prod_{j} q_{\mathbf{\phi}}(\mathbf{z}^{\text{(i)}})_{\text{t, j}}))
\end{align}
We can now use $\beta$ to increase or decrease the temporal independence of the factors we learn. This is equivalent to using a TC-VAE \cite{chen2018isolating} with a minimum $\beta$ of $1$ and a different estimation of the TC. 

\paragraph{Implementation}
\label{sec:experiments}
The algorithms are implemented using Pytorch \cite{paszke2017automatic} and are trained on an internal cluster using single NVIDIA GeForce 2800 and NVIDIA V100 GPUs, with a batch size of $8$, the Adam optimizer\cite{kingma2014adam}, a 1E-4 weight decay, a learning rate of 5E-3, $0.1$ epsilon, and $0.9$, $0.999$ as betas. Each instantiation of the algorithm takes about $3-4$ hours to train, based on the graphics card. We also reduce the learning rate when it plateaus using a scheduler, with a $0.95$ factor reduction on each plateau, patience of $6$ epochs, and a minimum learning rate of 1E-5. There is also specifically L2 norm regularization on the weight matrix between hidden states in the spatial decoder. The seed we use to train the models is $42$ and each model is trained for $150$ epochs, the last epoch is used for the evaluation and/or figures. All necessary code to download and preprocess the data, and run the model will be made publicly available after the double-blind review has concluded on GitHub. 

\paragraph{ICA null model}
Independent Component Analysis (ICA), has been used as a blind source separation to determine different sources of spatial or temporal signals that mix to form the measured signal. The algorithm maximizes the independence of these sources based on either spatial or temporal dissimilarities. ICA has long been used as a gold standard in all neural data, due to its ability to separate sources of neural activity, as well as separate non-neuronal activity, such as motion, respiration, and heartbeat effects. Over time, it has been established as the gold standard in separating spatio-temporal dynamics in EEG, ECOG, MEG, as well as in fMRI datasets \cite{calhoun2009review}. We, therefore, utilize ICA as a null model in order to compare our algorithm. The temporal independence results are compared to InfoMax ICA \cite{lee1999independent} with the same number of factors as our proposed model. The shortcomings of ICA are that, unlike PCA or other dimensionality techniques, the ICA vectors are unordered and sometimes need manual selection. Moreover, ICA vectors can be noisy for high dimensional data, and prior knowledge such as spatial or temporal constraints established in our spectral clustering cannot be trivially added to the algorithm.

\paragraph{Hyperparameter determination of Weight sharing \& temporal independence}
\label{sec:exp:weight-sharing}
The first experiment revolves around comparing weight sharing to a more na\"{i}ve implementation. The na\"{i}ve implementation uses two separate linear layers for each hemisphere, with an output of $128$ each, followed by an ELU activation \cite{clevert2015fast}. The two $128$-dimensional outputs are then concatenated and mapped to a $128$-dimensional vector with ELU activation \cite{clevert2015fast}, which is used as input to the temporal decoder. The decoder consists of an exact symmetric version of the encoder, with a Tanh activation after the final layer. This is similar to methods such as independent component analysis that associate a separate weight with each input dimension. We compare both functional and structural weight sharing with the baseline to make sure the reconstructions are at least as good as the baseline, but with a large reduction in parameters. The number of parameters for the functional and structural weight sharing is ~413k and ~417k, respectively, and  ~3.2M for the baseline method. This is roughly a 10x reduction in the number of parameters that scale with the number of input voxels. Given that the number of voxels in an fMRI volume is around 100-200k, which is 10-20x more than in this work (~11k), weight sharing is incredibly important to scale our approach up to larger input sizes. Furthermore, the baseline only has two non-linearities in the encoder and decoder, which means that the model can not learn functions as complex as deeper networks, such as our proposed model. This may limit the non-linearity of the factors it can learn, which can be an issue with more complex data types, such as resting-state fMRI data or multimodal fusion.

The experiments for the baseline, ICA, and other models are run with $16$ latent factors/components and a spatial encoder output size of $128$. The experiments are performed for $\beta = 0.0, 0.1, 0.25, 0.5, 0.75, 1.0, 2.0, 3.0, 4.0$, and $5.0$ to understand the effect of the $\beta$ hyperparameter on the performance of the model.

\section{Results}
\label{sec:results}
Our results show how well our model can correctly identify relevant latent factors from fMRI data. The first section discusses the performance of structural and functional weight sharing to its baseline. We show that the weight sharing we induce is effective and seems to even improve the reconstructions. The algorithm is also demonstrated to outperform the null ICA model for latent factor identification. Lastly, we show how these spatial maps localize to the motor homunculus, and use t-SNE \cite{van2008visualizing} to show a $2D$ view of the clustering of sub-tasks in the latent factor space.

\begin{figure}
    \centering
    \includegraphics[width=0.75\textwidth]{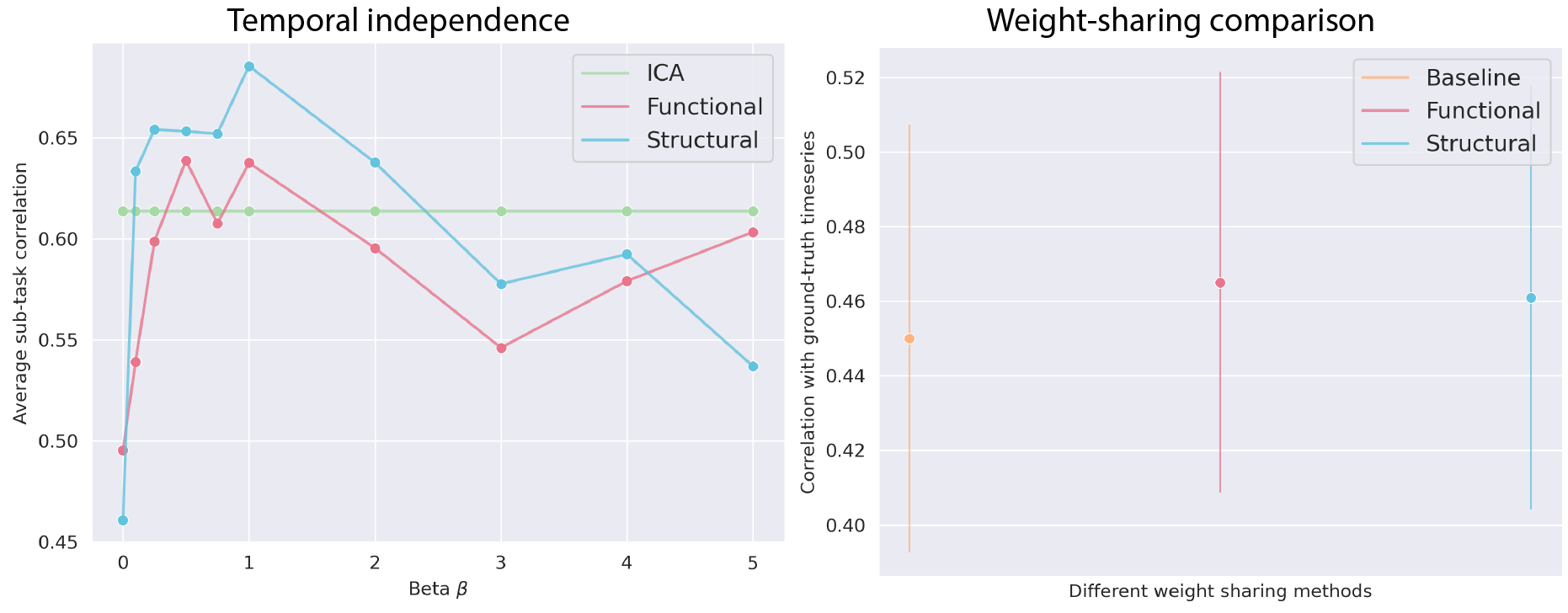}
    \caption{The left subfigure shows the average correlation of the factors corresponding to each sub-task. The temporal independence term in the loss function (Equation \ref{eq:final-bound}) clearly improves the solutions of the learned factors. The right subfigure shows the correlation between the ground-truth timeseries and the reconstructed timeseries. Functional and structural weight sharing outperforms the baseline on average, although both are within one standard deviation of the baseline (transparent vertical line). Importantly, however, the baseline has about 10x more parameters and performs little weight sharing.}
    \label{fig:correlation}
\end{figure}

\paragraph{Weight sharing \& temporal independence}
\label{sec:results:reconstruction}
The weight sharing results are shown in Figure \ref{fig:correlation} on the right. All models appear similar, with large standard deviations over the subjects in the unseen test set. On average, the functional weight sharing seems to be most performant, with structural weight sharing following closely behind. The baseline has the worst performance, although within a standard deviation of the other two models. Even so, this result indicates that neuroscientifically-backed weight sharing can lead to large reductions in the number of parameters and more efficient model architectures. Slightly better performance for the functional inductive bias is expected because the signals for those voxels are explicitly based on the strength of temporal correlations between the voxels. Generally, the correlation is fairly high, about $0.46$ on average, this seems to indicate that the model maintains the general trend of the data in its hidden state.

\begin{figure}
    \centering
    \includegraphics[width=\textwidth]{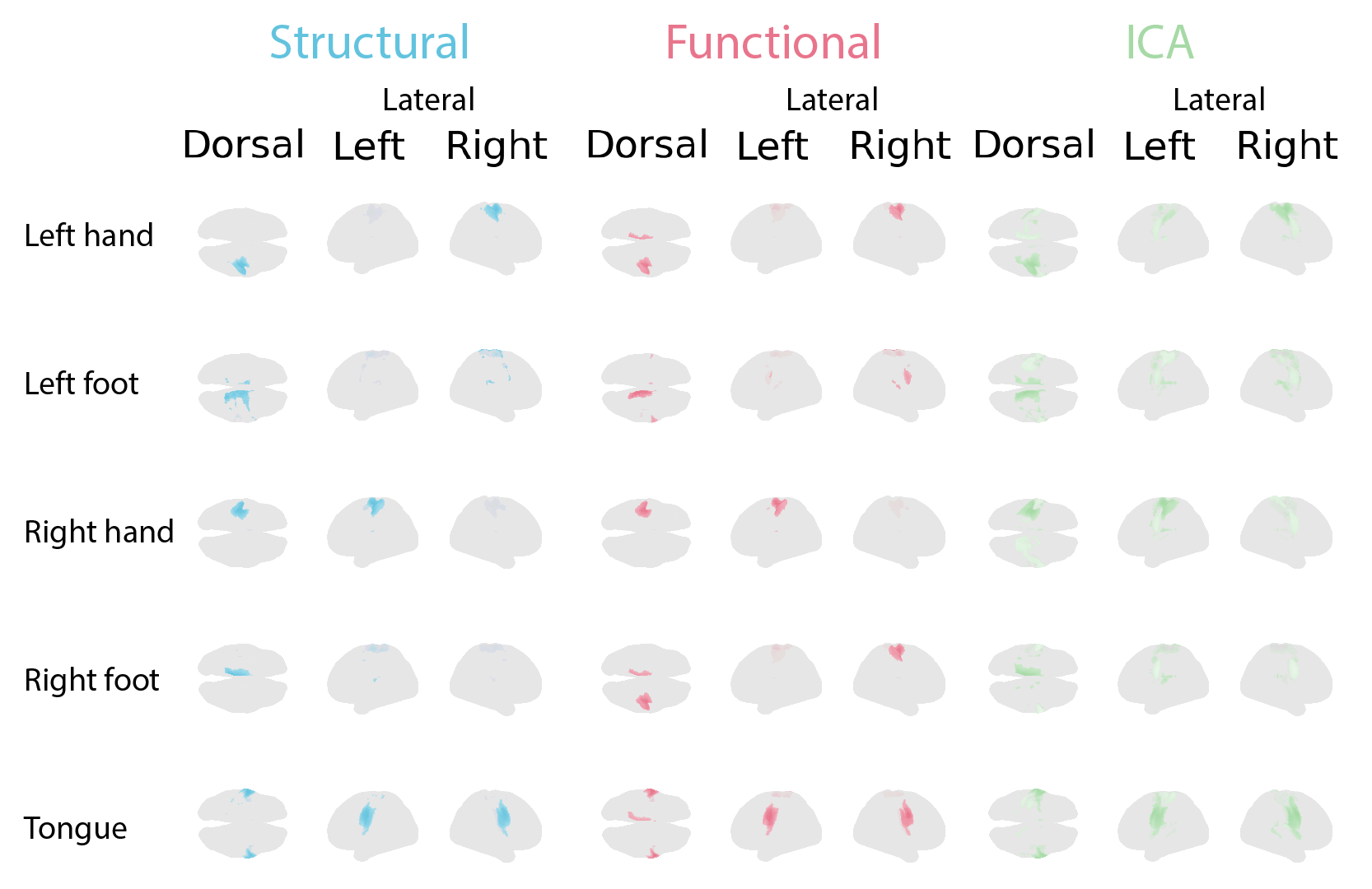}
    \caption{Each model's spatial maps corresponding to the sub-tasks (left), with structural weight sharing on the left, functional weight sharing in the middle, and ICA on the right. Note that for the dorsal view, the bottom hemisphere in the figure corresponds to the right hemisphere.}
    \label{fig:factors}
\end{figure}

The indication that the hidden state contains meaningful information regarding the spatio-temporal signal is supported by the high average sub-task correlations, shown in Figure \ref{fig:correlation} on the left. Both the functional weight sharing and the structural weight sharing outperform ICA for $\beta$s higher than $0.2$, but structural weight sharing clearly outperforms both methods. This result is a clear demonstration that our method is valuable, especially because our model is fully differentiable, non-linear, and can easily be extended to other data, or be combined with other modalities. To get some more insight into the spatial locations that correspond to the factors that have high correlations with the sub-tasks, we plot them for $\beta=1.0$ in Figure \ref{fig:factors}. The spatial maps for our model are created by interpolating each latent factor independently from $-3$ to $3$ in the latent space with $50$ steps and then taking the variance over those steps. The spatial maps are thus non-negative, whereas the ICA spatial maps can be negative. To deal with this, we use the sign of the correlation for each ICA factor with the sub-task it corresponds to and multiply the corresponding spatial map with its sign. All spatial maps are thresholded at $0.1$.

\begin{figure}
    \centering
    \includegraphics[width=\textwidth]{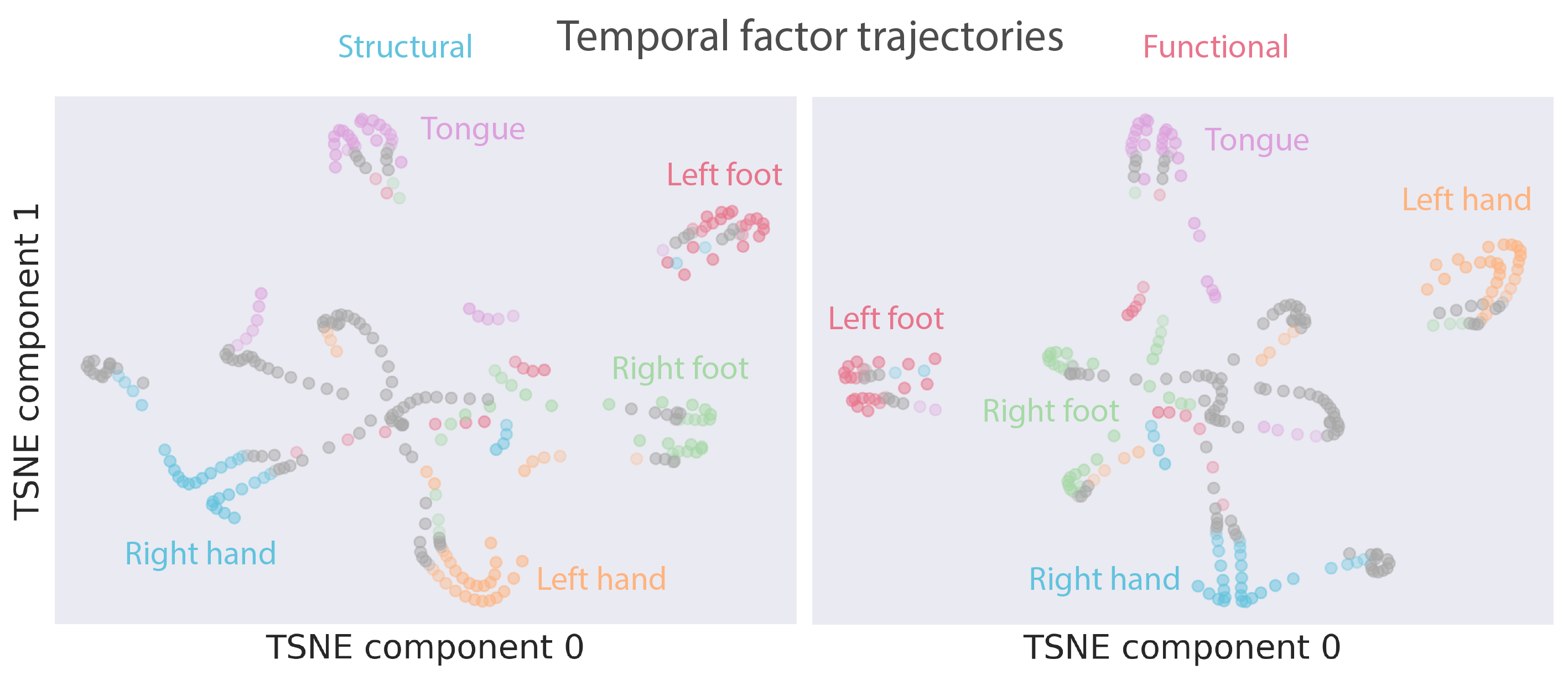}
    \caption{The average latent trajectory projected to $2D$ using t-SNE \cite{van2008visualizing} for the test set, for both the model with structural (left) and functional (right) inductive biases. Each point in the plot corresponds to a time point and the color corresponds to the task they are in, where gray corresponds to not being in a task. We expect a delay in the reaction due to the hemodynamic response, the first and last $5$ timesteps into a task have a gradually increasing and at the end decreasing opacity.}
    \label{fig:trajectory}
\end{figure}

\paragraph{Task dynamics}
\label{sec:results:task-dynamics}
To get an insight into the dynamics of the factors, we plot a $2D$ t-SNE \cite{van2008visualizing} projection of the average timeseries over the subjects in the unseen test set. Each point in Figure \ref{fig:trajectory} corresponds to a time point from that average timeseries and is colored based on which task it corresponds to, where gray points correspond to time points without a task. Since there is a delay in the BOLD response to a task, the first $5$ timepoints at the start of a task are made gradually more opaque, from $0.5$ to $1.0$, and the last $5$ timepoints are made gradually less opaque, from $1.0$ to $0.5$ for each task. We do not expect the first timepoints after the task starts to elicit a response, so some of the colored points may not be clustered together. The trajectories for both the structural (left) and functional (right) weight sharing are shown in Figure \ref{fig:trajectory}.

The trajectories in Figure \ref{fig:trajectory} show clear clusters for each sub-task, which are each performed twice in the timeseries. The same sub-task is not performed subsequently, making the clustering non-trivial. Interestingly, the structural weight sharing seems to create more separated clusters with the closeness between the left and right-hand clusters, and the left and right foot clusters. Furthermore, the shape of the trajectory is the same across the functional and structural weight sharing for most tasks. The right-hand task has a flipped T-shape, whereas the left-hand task is a semi-circular shape, for example. The exact shapes of the dynamics are an interesting future work.

\section{Discussion}
\label{sec:discussion}
The spatial maps in Figure \ref{fig:factors} correspond roughly to the functional motor homunculus. Namely, the spatial maps for the left and right hands are located superior and laterally on the left and right hemispheres, respectively. The locations of spatial maps for the left and right foot are located superior and more medially in the brain, on the left and right hemispheres, respectively. The spatial maps corresponding to the tongue are located inferior to the other sub-tasks, and laterally in both the right and left hemispheres. Given that our model learns these spatial maps over the whole dataset and that individual spatial maps can differ per subject, it is expected that the tongue spatial map occurs in both hemispheres. Another interesting finding is the difference between structural and functional spatial maps, namely that the structural inductive bias finds local regions, whereas the functional map sometimes finds regions more spread throughout the brain. This aligns with our expectations because spectral clusters for the functional inductive bias are based on temporal correlations, compared to the geodesic distance for the structural inductive bias. Additionally, both the structural and functional inductive bias find more localized regions that correspond more directly to the functional human motor homunculus, clearly indicating the usefulness of our model. The relationships between the timeseries of each factor and the sub-task may be linear in some cases, which would mean ICA is more appropriate. Given that our model learns those components and can learn non-linear components, our framework opens up a field of future work with non-linear fMRI components.

\paragraph{Limitations}
\label{sec:limitations}
One limitation of the model is that it has only been applied to the somatomotor cortex. This was done to have a good idea of ground-truth spatial and temporal factors we expect to find with our model. The somatomotor cortex is an extensively studied area and has largely been mapped out from a whole-brain perspective. However, it is important to test our model on larger input data in future work to make sure it holds up for whole-brain data. Furthermore, the SPM simulated hemodynamic response is not a perfect model for BOLD activation in the brain and we use a group surface to create the spectral graph clustering, instead of subject-based surfaces.

\paragraph{Broader impact}
\label{sec:broader}
The current model can have implications for surgical mapping, where functional connectivity based on ICA components is sometimes used. This model does require further and more extensive testing before it can be used in a clinical setting, however. The model's ability to learn non-linear factors can be both a positive and negative aspect of the model. The model can learn subject-specific factors that are not linearly related to group-based factors, as is common in ICA. This is important in a clinical setting, but could potentially lead to learning negative biases in the dataset.

\paragraph{Conclusion}
The model we propose in this work is a leap toward a fully-differentiable non-linear framework for whole-brain dynamic factor learning. We show that temporal independence is crucial to learning meaningful factors and our model outperforms ICA when the extra term that encourages temporal independence is added to the loss function. Our model can also comfortably scale to larger inputs with its novel weight sharing technique. In fact, weight sharing in our model does not degrade the reconstructions of the data under large dimensionality reduction (from $~11k$ voxels to $16$ factors) compared to a baseline. In future work, we want to apply this model to more tasks, larger input data, multiple modalities, and resting-state fMRI data. 

\paragraph{Acknowledgements}
Data were provided by the Human Connectome Project, WU-Minn Consortium (Principal Investigators: David Van Essen and Kamil Ugurbil; 1U54MH091657) funded by the 16 NIH Institutes and Centers that support the NIH Blueprint for Neuroscience Research; and by the McDonnell Center for Systems Neuroscience at Washington University
Other acknowledgments will be added once the double-blind review process has concluded. This material is based upon work supported by the National Science Foundation under Grant No. 2112455. Eloy Geenjaar was supported by the Georgia Tech/Emory NIH/NIBIB Training Program in Computational Neural-engineering (T32EB025816).
%%%%%%%%%%%%%%%%%%%%%%%%%%%%%%%%%%%%%%%%%%%%%%%%%%%%%%%%%%%%
\medskip

\printbibliography

%\input{checklist}

%\appendix

%\section{Appendix}

%Optionally include extra information (complete proofs, additional experiments and plots) in the appendix.
%This section will often be part of the supplemental material.

\end{document}